\documentclass[letterpaper, 10 pt, conference]{IEEEconf}
\pdfoutput=1
\IEEEoverridecommandlockouts

\usepackage{booktabs}
\usepackage{mathtools}

\usepackage{amssymb}
\usepackage{gensymb}
\usepackage{stmaryrd}
\usepackage{cite}
\usepackage{color}
\usepackage{multirow}
\usepackage{graphicx,times}
\usepackage{epstopdf}
\usepackage{indentfirst}
\usepackage{amsmath}
\usepackage{amsfonts}
\usepackage{txfonts}
\usepackage{mathrsfs}
\usepackage{theorem}
\usepackage{float}
\usepackage{subfigure}

\usepackage{verbatim}
\usepackage{hhline}     		
\usepackage{algorithmic}
\usepackage{algorithm}
\usepackage{array}
\usepackage{url}
\usepackage{bm}
\usepackage{xcolor}
\usepackage{microtype}
\usepackage{cleveref}
\usepackage{balance}

\crefname{section}{\S}{\S\S}
\Crefname{section}{\S}{\S\S}

\newcommand{\ie}{{\em i.e.}}
\newcommand{\eg}{{\em e.g.}}
\newcommand{\et}{{\em et al.}}

\def\spth{\textsuperscript{th}}

\title{Enhancing mmWave Radar Point Cloud via Visual-inertial Supervision}

\author{Cong Fan$^{1}$, Shengkai Zhang$^{1}$$^{*}$, Kezhong Liu$^{1}$, Shuai Wang$^{2}$, Zheng Yang$^3$, Wei Wang$^{4}$
\thanks{Authors$^{1}$ are with Wuhan University of Technology, Wuhan, China.
        {\tt\small \{congfan, shengkai, kzliu\}@whut.edu.cn}}%
\thanks{Author$^{2}$ is with Southeast University, Nanjing, China.
        {\tt\small shuaiwang@seu.edu.cn}}%
\thanks{Author$^{3}$ is with Tsinghua University, Beijing, China.
        {\tt\small hmilyyz@gmail.com}}%
\thanks{Author$^{4}$ is with Huazhong University of Science and Technology, Wuhan, China.
        {\tt\small weiwangw@hust.edu.cn}}%
\thanks{$^{*}$Corresponding author: Shengkai Zhang (\tt\small shengkai@whut.edu.cn).}
}

\begin{document}

\maketitle
\thispagestyle{empty}
\pagestyle{empty}

\begin{abstract}
Complementary to prevalent LiDAR and camera systems, millimeter-wave (mmWave) radar is robust to adverse weather conditions like fog, rainstorms, and blizzards but offers sparse point clouds. Current techniques enhance the point cloud by the supervision of LiDAR's data. However, high-performance LiDAR is notably expensive and is not commonly available on vehicles. This paper presents mmEMP, a supervised learning approach that enhances radar point clouds using a low-cost camera and an inertial measurement unit (IMU), enabling crowdsourcing training data from commercial vehicles. Bringing the visual-inertial (VI) supervision is challenging due to the spatial agnostic of dynamic objects. Moreover, spurious radar points from the curse of RF multipath make robots misunderstand the scene. mmEMP first devises a dynamic 3D reconstruction algorithm that restores the 3D positions of dynamic features. Then, we design a neural network that densifies radar data and eliminates spurious radar points. We build a new dataset in the real world. Extensive experiments show that mmEMP achieves competitive performance compared with the SOTA approach training by LiDAR's data. In addition, we use the enhanced point cloud to perform object detection, localization, and mapping to demonstrate mmEMP's effectiveness.
\end{abstract}

\section{Introduction}
\label{sec:intro}
The utilization of millimeter-wave (mmWave) radar has achieved significant prevalence in automotive and robotic applications. As the mmWave radio frequency (RF) is robust to perceive environments through small particles, \eg, smoke, fog, and snow, mmWave radar is complementary to optical sensors such as cameras and LiDAR~\cite{wang2023human, li2022pedestrian, cao2022cross, cai2022autoplace, gao2022dc, prabhakara2023high}. Recently, single-chip 4D mmWave radar~\cite{4d2023jiang, ding2023hidden, ding2022self} extends the sensing capability from 2D to 3D space. Although both 4D radar and 3D LiDAR provide point clouds to describe 3D environments, 4D radar is more favorable for autonomous vehicles as the manufacturing cost of 4D radar is one order of magnitude lower than that of 3D LiDAR (\$$550$ for Continental $77$ GHz ARS408 radar vs. \$$4000$ for Velodyne's $16$-line LiDAR).

The downside of mmWave radar is that its point cloud is two orders of magnitude sparser than a LiDAR's data due to the larger wavelength of RF~\cite{prabhakara2023high, dong2023gpsmirror} (compared with the nanometer-level wavelength of optical signals). The low spatial resolution comes from the RF's specular reflection and the low angular resolution. Current approaches adopt multi-sensor fusion to enhance the radar spatial resolution and enable various robot applications, \eg, object detection, localization, and mapping, in vision/LiDAR-crippled environments~\cite{ding2023hidden, lu2020milliego, lu2020see, cheng2022novel, prabhakara2023high}. However, these approaches must collect training data from expensive 3D LiDARs, typically equipped on a few specialized vehicles. The limited number of specialized vehicles will limit the scale of training data, being prone to the long tail of autonomous vehicle perception.

\begin{figure}[t!]
  \centering
  \includegraphics[width=3.2in]{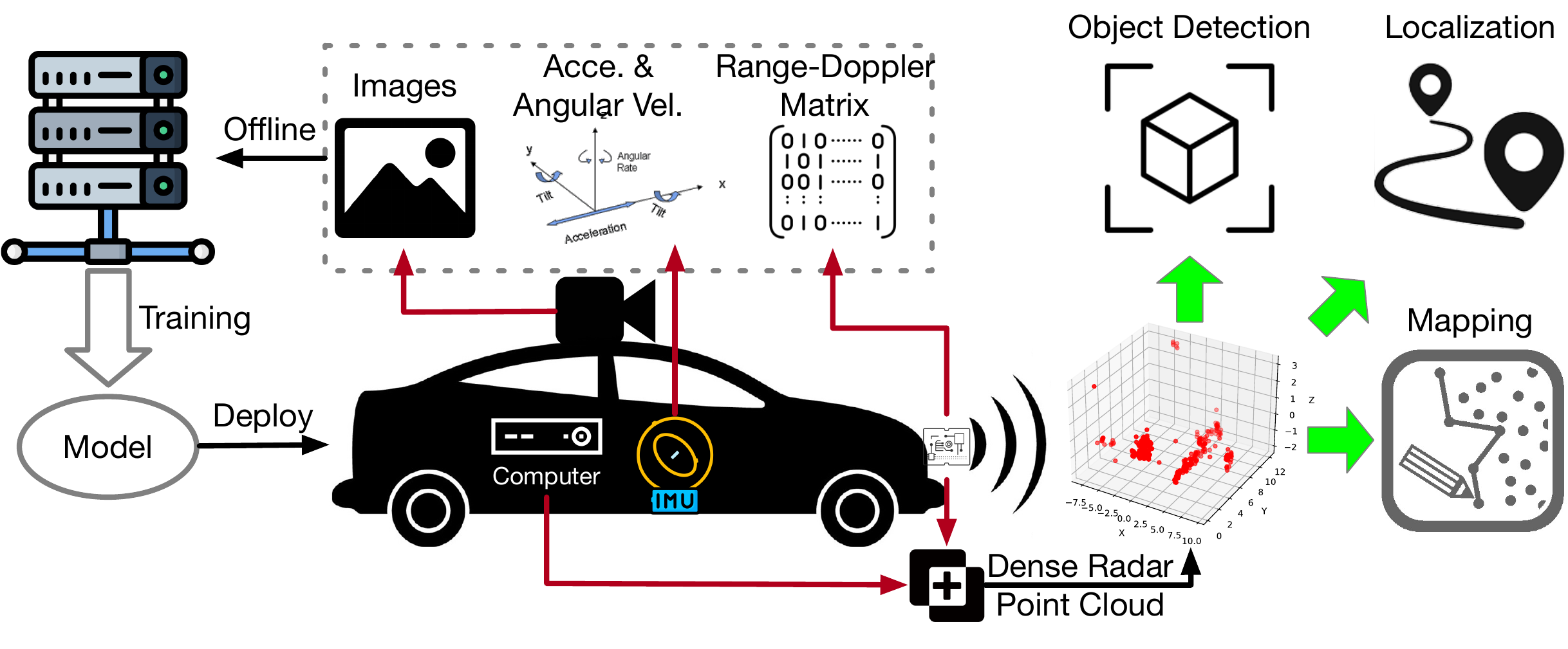}
  \caption{mmEMP takes images, inertial measurements, and range-Doppler matrices (RDMs) to train a model for enhancing radar point clouds. In the test, the vehicle uses the enhanced point clouds to improve various applications, \eg, object detection, localization, and mapping.}
  \label{fig:toy}
  \vspace{-6mm}
\end{figure}

We propose mmEMP, a supervised learning approach that enhances the point cloud of a mmWave radar using camera images and inertial measurements, as shown in Fig.~\ref{fig:toy}. Since inertial measurement units (IMUs) and cameras are low-cost and commonly installed on commercial vehicles like Tesla, mmEMP goes towards crowdsourcing the training data from intelligent vehicles. Typically, a LiDAR's dense point cloud directly provides 3D masks of objects in the scene. It is a solid supervision to guide the spatial information of real objects {\em w.r.t.} the radar data. In contrast, a visual image only provides 2D pixels, lacking depth information. Nevertheless, the state-of-the-art (SOTA) visual-inertial (VI) simultaneous localization and mapping (SLAM) techniques~\cite{qin2018vins, campos2021orb} can reconstruct the 3D scene by dense visual features, holding the opportunity that provides LiDAR-like supervision.

In realizing mmEMP, we encounter two challenges. First, the fundamental of VI 3D reconstruction is epipolar geometry, assuming a fixed point in the world projects on multiple camera frames. A moving point breaks this assumption in that the projections on different frames correspond to points in its moving trajectory. The reconstructed result is the intersection of the moving point {\em w.r.t.} the camera center across multiple frames (refer to Fig.~\ref{fig:dynamic_geometry}). Second, there exist spurious points that represent no real object due to the multipath reflection of RF signals. Densifying these spurious points will make robots misunderstand the world.


mmEMP addresses the above challenges by two modules. 

{\em First}, we devise a dynamic 3D reconstruction algorithm that restores the 3D positions of moving features. The movement breaks the epipolar constraint. Our algorithm builds upon the critical observation that all points on a rigid object share the identical translation between camera frames, allowing us to formulate the dynamic 3D reconstruction into a non-linear least square problem. 

{\em Second}, we design a neural network pipeline that takes the VI 3D reconstruction to densify radar point clouds and eliminate spurious points. We eliminate spurious points by checking the spatial stability across consecutive frames. It requires the corresponding rigid transformations to transform coordinate frames. Thus, the neural network first estimates rigid transformation. Then we refine the point cloud using a spatial stability checking algorithm.


{\bf Contributions}. mmEMP makes three contributions: 
\begin{itemize}
	\item We propose a dynamic VI 3D reconstruction algorithm that restores the 3D positions of dynamic visual features.
	\item We design a neural network pipeline that takes VI data and radar RDMs to enhance radar point clouds, estimate the vehicle's pose, and eliminate spurious radar points.
	\item We build a large dataset of radar RDMs, camera images, and IMU sequences along with open-source code\footnote{The open-source code and dataset of mmEMP is available at \url{https://github.com/bella-jy/mmEMP}.}.
\end{itemize}






\section{Related Work}
\label{sec:related}
{\bf Radar point cloud enhancement}. Currently, point cloud enhancement approaches for mmWave radar either leverage advanced signal processing techniques, \eg, Synthetic Aperture Radar (SAR) imaging~\cite{qian20203d, yamada2017high, ghasr2016wideband, watts20162d, guan2020through, luo2021single}, or fuse with other sensors, \eg, camera and LiDAR~\cite{yang2022ralibev, wang2023bi, kim2020grif, nabati2021centerfusion, hwang2022cramnet, wang2021rodnet, huang2021cross, nabati2020radar, qian2021robust, ding2023hidden, lu2020milliego, lu2020see, cheng2022novel, prabhakara2023high}. SAR imaging requires a vehicle to move along a specific trajectory precisely. For example, Qian~\et~\cite{qian20203d} exploited the natural linear motion of vehicle radar to improve sensing resolution. Thus, the radar perception will be inferior when an automobile stops at an intersection or turns sharply.

On the other hand, enhancing mmWave radar perception by fusing with other sensors has been attractive. Among them, some take radars to improve the perception of LiDARs or cameras to perform better in challenging environments~\cite{yang2022ralibev, wang2023bi, kim2020grif, nabati2021centerfusion, hwang2022cramnet, wang2021rodnet, huang2021cross, nabati2020radar, qian2021robust}. The radar in these works cannot work alone. On the contrary, some other works try to enhance the spatial resolution of radars by the supervision of LiDAR's dense point clouds~\cite{ding2023hidden, lu2020milliego, lu2020see, cheng2022novel, prabhakara2023high}. Although the training requires data from multiple sensors, the radar alone performs robustly in various robot applications. However, costly LiDAR prevents massive deployment and thus makes the system collect training data inefficient. 

{\bf Dynamic visual feature tracking}. Dynamic features have always been a challenge in SLAM. Currently, most SLAM techniques~\cite{qin2018vins, song2022dynavins, fan2019dynamic, canovas2020speed, dai2020rgb} reject dynamic features to ensure pose estimation accuracy. Qiu~\et~\cite{qiu2019tracking} simultaneously tracked the camera poses and dynamic objects by motion correlation analysis. They have proved that this problem is partially observable. Our dynamic 3D reconstruction is observable since we assume that dynamic features are outliers, so the camera pose can be first estimated. Moreover, we only need to recover the 3D positions of dynamic features rather than an object's 6-DoF pose. Ren~\et~\cite{ren2022visual} and Xu~\et~\cite{xu2019mid} focused on object-level tracking by learning-based semantic segmentation. In contrast, we focus on feature-level tracking for our point cloud enhancement.


\section{System Design of mmEMP}
\label{sec:design}
\subsection{Dynamic Visual-Inertial 3D Reconstruction}
\label{subsec:dynamic}
mmEMP aims to enhance the spatial resolution of mmWave radar and generate dense point clouds using a low-cost VI sensor suite. The opportunity for this idea comes from the 3D reconstruction of VI SLAM~\cite{qin2018vins, campos2021orb}. The reconstructed 3D positions of dense visual features may provide LiDAR-like guidance for radar point cloud generation. 

\begin{figure}[t!]
    \centering
    \shortstack{
            \includegraphics[width=0.235\textwidth]{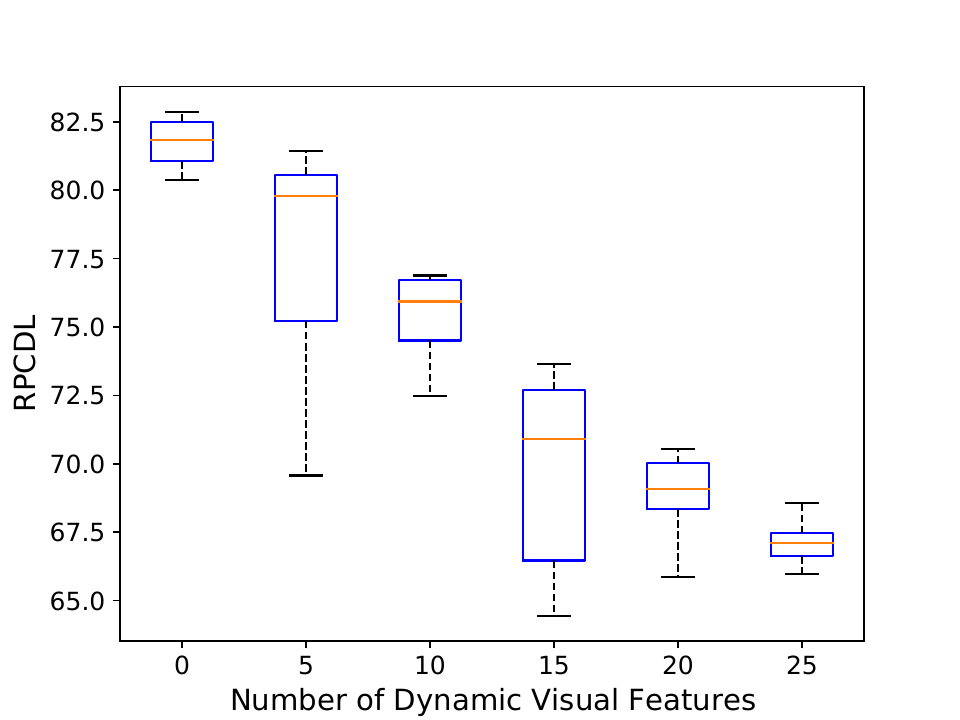}\\
            {\footnotesize (a) Statistical RPCDL.}
    }
    \shortstack{
            \includegraphics[width=0.235\textwidth]{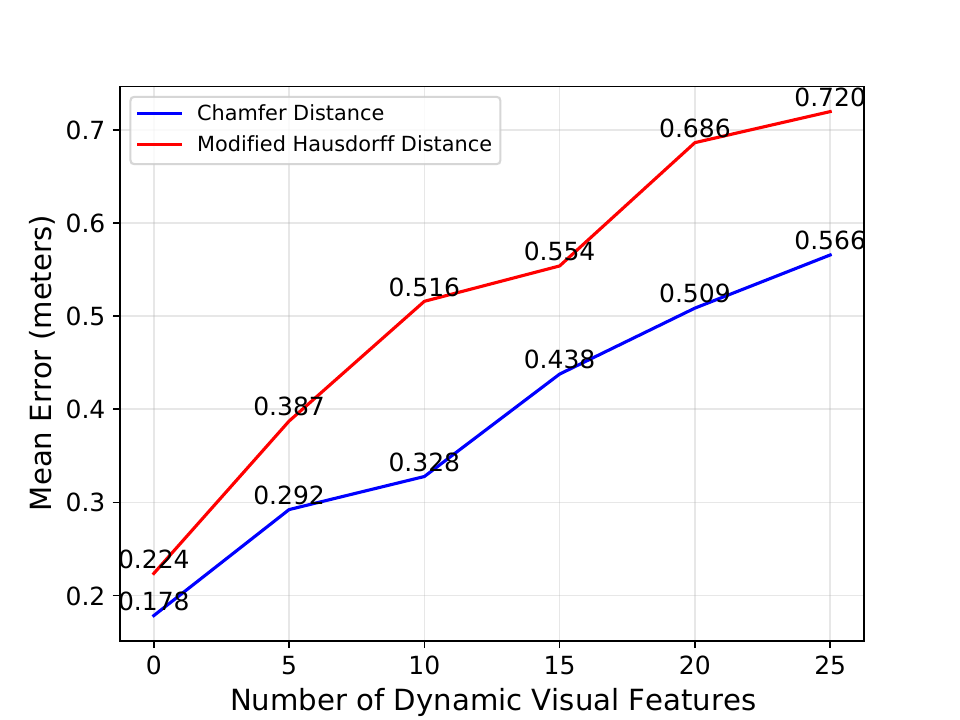}\\
            {\footnotesize (b) Chamfer and Hausdorff distances.}
    }
    \caption{A preliminary study shows that dynamic visual features with wrong 3D positions significantly degenerate the performance of point cloud generation.}
    \label{fig:study}
    \vspace{-5mm}
\end{figure}


{\bf Preliminary study}. To verify this idea, we perform the VI 3D reconstruction by VINS~\cite{qin2018vins} and input the 3D positions of visual features into the SOTA radar-based point cloud generation method~\cite{cheng2022novel}. Extensive experiments show that the moving visual features in the scene, \ie, dynamic features, cause performance degeneration. We conduct the experiments by controlling the number of dynamic features from $0$ to $30$ with a step of $5$. We conduct ten trials for each setting and plot the statistical result of the radar point cloud density level (RPCDL)~\cite{cheng2022novel} in Fig.~\ref{fig:study}(a). In addition, we calculate the mean Chamfer~\cite{chamfer} and Hausdorff~\cite{hausdorff} distances to measure the point cloud similarity, as shown in Fig.~\ref{fig:study}(b). 

When we select the scene without moving objects, the RPCDL (the larger, the better) is similar to~\cite{cheng2022novel}. However, the performance exhibits a remarkable decrease if we increase the number of dynamic features. In contrast, the mean Chamfer and Hausdorff distances increase along with more dynamic features, meaning that the generated point cloud differs from the ground truth point cloud obtained by the depth map from Intel Realsense D435. The reason is that VI SLAM techniques cannot restore dynamic features' 3D positions because they break the epipolar constraint for bundle adjustment. The VI 3D reconstruction cannot correctly capture all dynamic objects in the scene. Next, we elaborate on the problem of dynamic 3D reconstruction.

\begin{figure}
	\centering
	\begin{minipage}[b]{0.23\textwidth}\centering
		\center
		\includegraphics[width=1\textwidth]{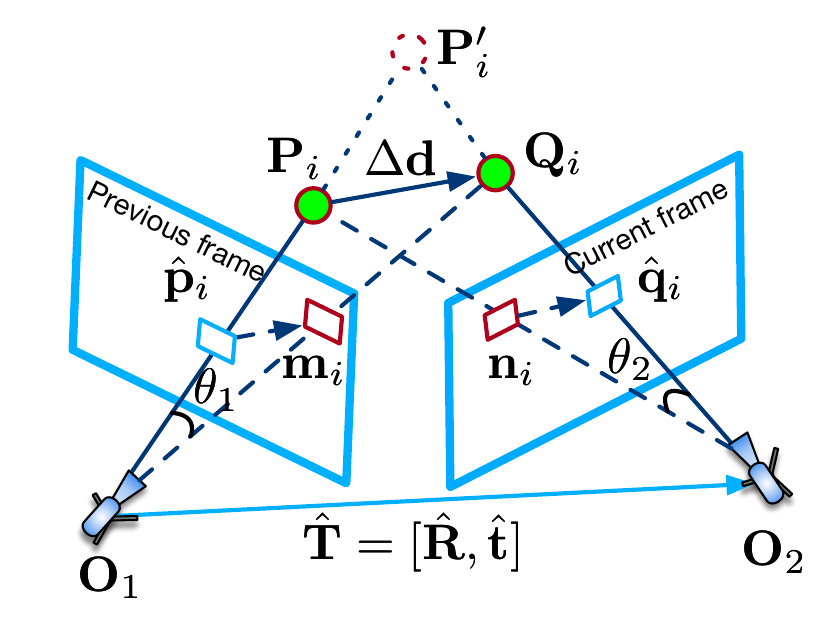}\vspace{-0.3cm}
		\caption{The geometry of a dynamic feature between two camera frames.} 
		\label{fig:dynamic_geometry}
	\end{minipage}
	\hspace{0.1cm}
	\begin{minipage}[b]{0.23\textwidth}\centering
		\center
		\includegraphics[width=1\textwidth]{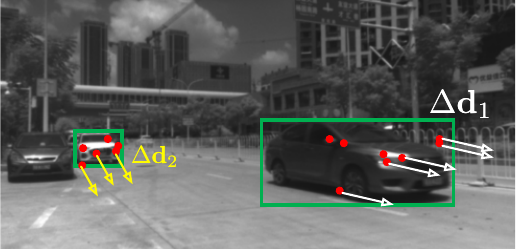}\vspace{-0.3cm}
		\caption{Dynamic features on a rigid object share the same translation.} 
		\label{fig:trans}
	\end{minipage}
	\vspace{-0.4cm}
\end{figure}

{\bf Problem formulation}. Fig.~\ref{fig:dynamic_geometry} illustrates the geometry of a dynamic feature between two consecutive frames. Consider a moving point $i$ observed on the previous frame at pixel\footnote{In our formulation, we mark $\hat{(\cdot)}$ as known variables or constants. $\|\cdot\|$ denotes the $L^2$ norm unless otherwise specified. We also express the 2D pixel as its homogeneous coordinate.} $\hat{\mathbf{p}}_i = [\hat{u}_{pi}, \hat{v}_{pi}, 1]^\top$ and on the current frame at pixel $\hat{\mathbf{q}}_i = [\hat{u}_{qi}, \hat{v}_{qi}, 1]^\top$, features $\hat{\mathbf{p}}_i$ and $\hat{\mathbf{p}}_i$ are paired by optical flow. The camera pose $\hat{\mathbf{T}} = [\hat{\mathbf{R}}, \hat{\mathbf{t}}]$, where $\hat{\mathbf{R}} \in \text{SO}(3)$ and $\hat{\mathbf{t}} \in \mathbb{R}^3$, is known by VI SLAM~\cite{qin2018vins, campos2021orb} as long as dynamic features are substantially fewer than background features, which is usually true in practice. The outlier rejection technique (RANSAC~\cite{fischler1981random}) of VI SLAM makes the robot pose estimation robust to these dynamic features. 

Our goal is to calculate the 3D position $\mathbf{P}_i\in\mathbb{R}^3$ {\em w.r.t.} the previous frame and the translation $\Delta\mathbf{d}\in\mathbb{R}^3$ when it is observed on the current frame. The epipolar constraint requires a fixed point in the scene, while the dynamic point has a translation when observed in the second frame. Existing techniques calculate the intersection $\mathbf{P}_i^\prime$ of the moving point {\em w.r.t.} the camera center.  


The translation makes the point re-projection no longer exists. Instead, both frames share the projection of the translation. Consider $\mathbf{O}_1 = (0, 0, 0)$, from the cosine law, we have the following equations (refer to Fig.~\ref{fig:dynamic_geometry}):
\begin{equation}
	\|\mathbf{P}_i\|^2 + \|\mathbf{Q}_i\|^2 - 2\|\mathbf{P}_i\|\cdot\|\mathbf{Q}_i\| \cos\theta_1 - \|\Delta\mathbf{d}\|^2 = 0,
	\label{eqn:left}
\end{equation}
\begin{equation}
	\|\mathbf{P}_i-\hat{\mathbf{t}}\|^2 + \|\mathbf{Q}_i-\hat{\mathbf{t}}\|^2 - 2\|\mathbf{P}_i-\hat{\mathbf{t}}\|\cdot\|\mathbf{Q}_i-\hat{\mathbf{t}}\| \cos\theta_2 - \|\Delta\mathbf{d}\|^2 = 0,
	\label{eqn:right}
\end{equation}
where $\mathbf{Q}_i = \mathbf{P}_i + \Delta\mathbf{d}$. 

To explicitly express $\cos\theta_1$ and $\cos\theta_2$ {\em w.r.t.} $\mathbf{P}_i$ and $\Delta\mathbf{d}$, we define the pseudo-projections of the moving point $i$. $\mathbf{m}_i \in \mathbb{R}^3$ denotes the pseudo-pixel homogeneous coordinate of point $\mathbf{Q}_i$ in the previous frame. $\mathbf{n}_i \in \mathbb{R}^3$ is the pseudo-projection homogeneous coordinate of point $\mathbf{P}_i$ in the current frame. Then, we have
\begin{equation}
	\cos\theta_1 = \frac{\hat{\mathbf{p}}_i \cdot \mathbf{m}_i}{\|\hat{\mathbf{p}}_i\| \cdot \|\mathbf{m}_i\|}, \;\; \cos\theta_2 = \frac{\hat{\mathbf{q}}_i \cdot \mathbf{n}_i}{\|\hat{\mathbf{q}}_i\| \cdot \|\mathbf{n}_i\|}.
\end{equation}
Next, we express the pseudo-projections $\mathbf{m}_i$ and $\mathbf{n}_i$ {\em w.r.t.} the target unknowns $\mathbf{P}_i$ and $\Delta\mathbf{d}$.

Given the camera intrinsic matrix $\hat{\mathbf{K}} = \begin{bmatrix}
	\hat{f}_x & 0 & \hat{c}_x \\
	0 & \hat{f}_y & \hat{c}_y \\
	0 & 0 & 1
\end{bmatrix}$ and $\mathbf{P}_i = [p_{ix},\; p_{iy},\; p_{iz}]^\top$, $\Delta\mathbf{d} = [\Delta d_x,\; \Delta d_y,\; \Delta d_z]^\top$, the pinhole imaging model gives: 
\begin{equation}
	\mathbf{m}_i = \begin{bmatrix}
		\hat{f}_x\frac{p_{ix}+\Delta d_x}{p_{iz}+\Delta d_z} + \hat{c}_x & \hat{f}_y\frac{p_{iy}+\Delta d_y}{p_{iz}+\Delta d_z} + \hat{c}_y & 1
	\end{bmatrix}^\top.
\end{equation}
In addition, the imaging model also gives a measurement model,
\begin{equation}
	\hat{\mathbf{p}}_i = \begin{bmatrix}
		\hat{f}_x\frac{p_{ix}}{p_{iz}} + \hat{c}_x & \hat{f}_y\frac{p_{iy}}{p_{iz}} + \hat{c}_y & 1
	\end{bmatrix}^\top.
	\label{eqn:m1}
\end{equation}

We now define the homogeneous coordinate of point $i$ in the previous frame as $\tilde{\mathbf{P}}_i = [\mathbf{P}_i; 1]$. In addition, we re-parameterize the known pose $\hat{\mathbf{T}} = \begin{bmatrix}
 	\hat{\mathbf{R}}, \hat{\mathbf{t}}
 \end{bmatrix} = \begin{bmatrix}
 	\hat{\mathbf{t}}_1 & \hat{\mathbf{t}}_2 & \hat{\mathbf{t}}_3
 \end{bmatrix}^\top$, where $\hat{\mathbf{t}}_j \in \mathbb{R}^{1\times 4}$, $j = \{1,\; 2,\; 3\}$, denotes $j$\spth row of $\hat{\mathbf{T}}$. Then the image projection theory gives:
 \begin{equation}
 	s\cdot \mathbf{n}_i = \begin{bmatrix}
 		\hat{\mathbf{t}}_1 & \hat{\mathbf{t}}_2 & \hat{\mathbf{t}}_3
 	\end{bmatrix}^\top\tilde{\mathbf{P}}_i,
 \end{equation}
 where $s$ denotes the scale variable. Eliminating $s$ by the last row of the above equation writes 
 \begin{equation}
 	\mathbf{n}_i = \begin{bmatrix}
 		\frac{\hat{\mathbf{t}}_1\tilde{\mathbf{P}}_i}{\hat{\mathbf{t}}_3\tilde{\mathbf{P}}_i} & \frac{\hat{\mathbf{t}}_2\tilde{\mathbf{P}}_i}{\hat{\mathbf{t}}_3\tilde{\mathbf{P}}_i} & 1
 	\end{bmatrix}^\top.
 \end{equation}
 Similarly, we have another measurement model with pixel $\hat{\mathbf{q}}_i$
 \begin{equation}
 	\hat{\mathbf{q}}_i = \begin{bmatrix}
 		\frac{\hat{\mathbf{t}}_1\left(\tilde{\mathbf{P}}_i+\Delta\tilde{\mathbf{d}}\right)}{\hat{\mathbf{t}}_3\left(\tilde{\mathbf{P}}_i+\Delta\tilde{\mathbf{d}}\right)} & \frac{\hat{\mathbf{t}}_2\left(\tilde{\mathbf{P}}_i+\Delta\tilde{\mathbf{d}}\right)}{\hat{\mathbf{t}}_3\left(\tilde{\mathbf{P}}_i+\Delta\tilde{\mathbf{d}}\right)} & 1
 	\end{bmatrix}^\top.
 	\label{eqn:m2}
 \end{equation} 

 At this stage, all terms in Eqn.~\eqref{eqn:left} and~\eqref{eqn:right} are expressed {\em w.r.t.} the target unknowns $\mathbf{P}_i$ and $\Delta\mathbf{d}$. 
 
 Combing Eqn.~\eqref{eqn:left}, \eqref{eqn:right}, \eqref{eqn:m1}, and \eqref{eqn:m2} writes the overall measurement model
 \begin{equation}
 	\hat{\mathbf{z}}_i = \begin{bmatrix}
 		0 & 0 & \hat{u}_{pi} & \hat{v}_{pi} & \hat{u}_{qi} & \hat{v}_{qi}
 	\end{bmatrix}^\top = F\left(\mathbf{P}_i,\; \Delta\mathbf{d}\right),
 	\label{eqn:meas}
 \end{equation}
 where $F\left(\mathbf{P}_i,\; \Delta\mathbf{d}\right)$ denotes the stacked vector of $6$ expressions {\em w.r.t.} unknown $\mathbf{P}_i$ and $\Delta\mathbf{d}$. Apparently, we cannot find the unique solution from Eqn.~\eqref{eqn:meas} due to its non-linearity.
 
 Our key observation is that dynamic features on a rigid object share the same translation $\Delta\mathbf{d}$ between camera frames, as shown in Fig.~\ref{fig:trans}. One more dynamic feature incurs $3$ more variables, \ie, its 3D position, but adds $6$ more measurement equations. Moreover, the SOTA image segmentation tool~\cite{kirillov2023segany} can separate the features of different rigid objects.
 
 Thus, we formulate the dynamic 3D reconstruction as a non-linear least square problem. We assume $N$ dynamic features on a rigid object. The unknown vector is $\mathbf{X} = \left(\mathbf{P}_1, \mathbf{P}_2, \cdots, \mathbf{P}_{N}, \Delta\mathbf{d}\right)$. Then, we have the following optimization problem
 \begin{equation}
 	\mathbf{X}^* = \arg\min_{\mathbf{X}} \frac{1}{2}\sum_{i=1}^N \left\|\mathbf{e}_i\left(\mathbf{X}\right)\right\|^2,
 \end{equation}
 where $\mathbf{e}_i(\mathbf{X}) = \hat{\mathbf{z}}_i - F\left(\mathbf{P}_i,\; \Delta\mathbf{d}\right)$. As long as $N\geq 2$ for a rigid object, the features' 3D positions and the translation can be recovered. We use Ceres Solver~\cite{ceres-solver}, an open-source C++ library, to solve this non-linear optimization problem.
 


\subsection{Point Cloud Generation and Refinement}
\begin{figure*}[t!]
  \centering
  \includegraphics[width=7in]{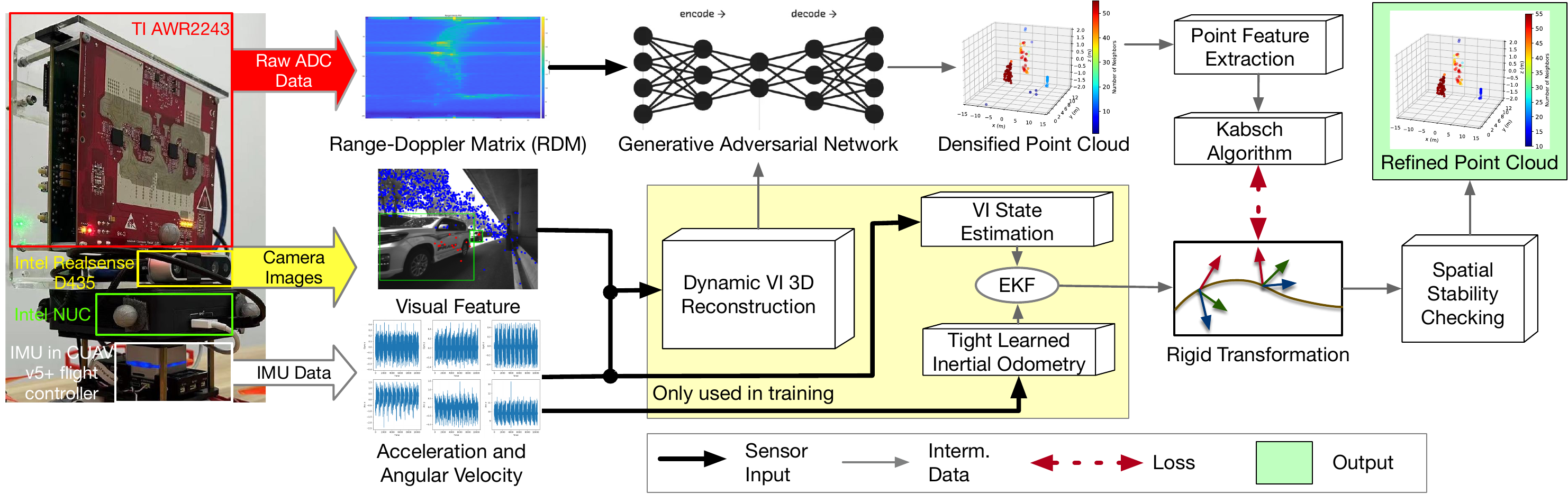}
  \caption{An overview of our data processing pipeline. The measurements from the visual-inertial sensor suit are only used in training (modules in the yellow box) so that mmEMP works fine in adverse weather conditions.}
  \label{fig:pipeline}
\end{figure*}


The point cloud generation using the supervision of 3D visual point clouds obtained in \S~\ref{subsec:dynamic} is similar to~\cite{cheng2022novel}. Our technical contribution lies in the point cloud refinement. We refine the point cloud by eliminating spurious points via inferring the vehicle's drift-free rigid transformations and checking the point spatial stability. The data processing pipeline is shown in Fig.~\ref{fig:pipeline}.


{\bf Point cloud generation}. We input raw RDMs and output dense point clouds with the supervision of a ground-truth label matrix {\em w.r.t.} real objects from our dynamic VI 3D reconstruction. In particular, we apply a generative adversarial network~\cite{luc2016semantic} that consists of a generator and a discriminator.  Since the number of real target cells in the RDM is a minority, we use the focal loss~\cite{lin2017focal} to deal with the sample imbalance problem. For more network details, please refer to~\cite{cheng2022novel}. 

{\bf Point cloud refinement}. It is well known that RF multipath reflections cause spurious points~\cite{bansal2020pointillism, lu2020see}. Our idea to remove such spurious points is to leverage spatial stability. Intuitively, points on real objects show stable appearances in adjacent frames. On the contrary, spurious points are the mirror points of real objects {\em w.r.t.} the receiving path of multipath reflections. Their appearances show a random spatial pattern over time. Thus, if we superimpose multiple frames of point clouds {\em w.r.t.} the initial frame, points on real objects should see many more neighbors than spurious points.

\begin{figure}[t!]
  \centering
  \includegraphics[width=3in]{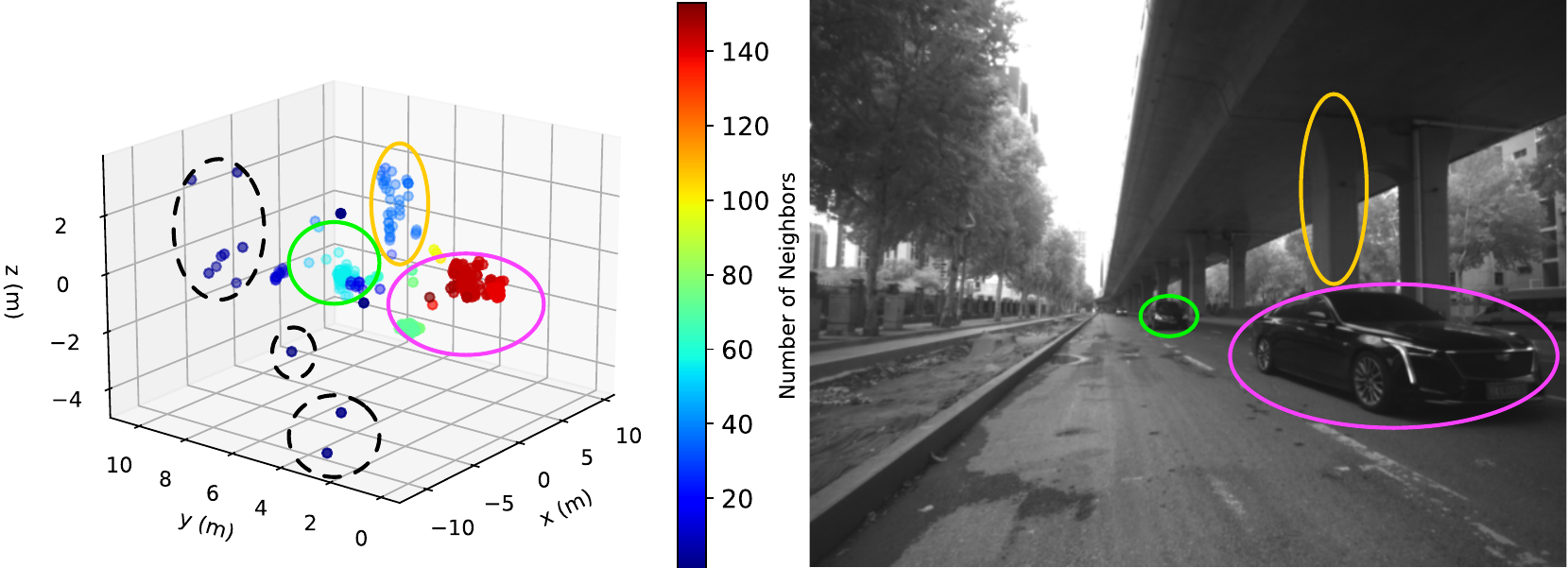}
  \caption{The naive spatial stability checking with known rigid transformations over a few adjacent frames. Solid circles are colored for different objects. Dotted dark circles highlight the spurious points.}
  \vspace{-6mm}
  \label{fig:spurious}
\end{figure}

To verify our idea, we conduct experiments in the outdoor scene with ground-truth rigid transformations provided by VINS~\cite{qin2018vins}. Fig.~\ref{fig:spurious} shows the preliminary result. We color points with different numbers of neighbors. For ease of verification, we stack $5$ adjacent frames of point clouds and set a distance threshold ($0.5$ m) to define a point's neighborhood. The result highlights the spurious points with dark blue colors in the scene. Some are positioned below the ground with a height of $-4.2$ meters. The result indicates that this approach is promising to identify spurious points. 

In practice, mmEMP aims to work in adverse weather where VI SLAM fails. Moreover, fixing a distance threshold to define the neighborhood will result in false positives of spurious points. Thus, we have the following designs.

{\em 1) VI-supervised rigid transformation learning.} We obtain reliable rigid transformations with radar point clouds through a supervised learning framework. The supervision of rigid transformation is from VI state estimation~\cite{qin2018vins, campos2021orb}. However, VI state estimations require loop closure to optimize its temporal drift. To mitigate the drift, we leverage the SOTA inertial-only state estimator, tight learned inertial odometry (TLIO)~\cite{liu2020tlio}, to be an independent source of rigid transformation. The inertial-only state estimation is enabled by extracting the motion pattern hidden in a segment of IMU measurements through a deep neural network. We fuse the rigid transformation of the VI state estimator~\cite{qin2018vins} with that of TLIO~\cite{liu2020tlio} by applying the extended Kalman filter (EKF) to obtain pseudo-ground-truth rigid transformations $\mathbf{T}_\text{truth}$ between radar frames (refer to Fig.~\ref{fig:pipeline}). 

We then use $\mathbf{T}_\text{truth}$ to supervise the inference of rigid transformations from radar point clouds. We first input the densified point cloud into the SOTA point feature extraction network~\cite{ding2023hidden}. The network gives a scene flow prediction $\hat{\mathbf{F}}=\{\hat{\mathbf{f}}_i \in \mathbb{R}^3\}_{i=1}^{M}$, where $M$ denotes the number of points in a radar frame, and a moving probability map $\hat{\mathbf{M}}=\{\hat{m}_i\in [0, 1]\}_{i=1}^{M}$, where $\hat{m}_i \geq 0.5$ indicates point $i$ is moving. Based on $\hat{\mathbf{F}}$ and $\hat{\mathbf{M}}$, we then infer the radar rigid transformation $\hat{\mathbf{T}} \in \text{SE}(3)$ by the Kabsch algorithm~\cite{kabsch1976solution}.

We aim to adjust the inferred radar transformation $\hat{\mathbf{T}}$ to approach the pseudo-ground-truth transformation $\mathbf{T}$ by a loss function. Consider two consecutive radar frames, namely, the previous frame $\mathcal{F}_p$ and the current frame $\mathcal{F}_c$, and the rigid transformation $\mathbf{T}_p^c = \begin{bmatrix}
	\mathbf{R}_p^c & \mathbf{t}_p^c \\
	\mathbf{0} & 1
\end{bmatrix}$ from $\mathcal{F}_p$ to $\mathcal{F}_c$. Then we can transform a point $\mathbf{f}_{pi}$ in $\mathcal{F}_p$ into $\mathcal{F}_c$ as $\mathbf{f}_{ci} =\mathbf{R}_p^c \mathbf{f}_{pi} + \mathbf{t}_p^c$. Thus, we can write the loss function of our supervised learning as
\begin{equation}
	L_\text{trans.} = \frac{1}{M}\sum_{i=1}^M\left\| \left(\mathbf{R}^\top\hat{\mathbf{R}} - \mathbf{I}_3\right)\mathbf{f}_{pi} + \mathbf{t} - \hat{\mathbf{t}} \right\|,
\end{equation}
where $\left(\mathbf{R}, \; \mathbf{t}\right)$ and $\left(\hat{\mathbf{R}}, \; \hat{\mathbf{t}}\right)$ are the rotation and translation components in $\mathbf{T}$ and $\hat{\mathbf{T}}$, respectively.

{\em 2) Velocity-adaptive spatial stability checking.} Through extensive experiments on the naive algorithm for idea verification, we realize there is no fixed threshold to determine a point's neighborhood for spatial stability checking. Intuitively, if a vehicle travels faster, the spatial distribution of radar points across multiple frames will be larger. Thus, we design a spatial stability checking algorithm shown in Algorithm~\ref{alg:spatial_check}.

For each point cloud frame, we first compute the relative velocities of background and dynamic points {\em w.r.t.} the current ($0$\spth) frame. The velocity of background points is the vehicle's (Lines $4-5$). However, dynamic points' relative velocities depend on the moving speeds of rigid objects. We compute this using the estimated translation by our dynamic 3D reconstruction in \S~\ref{subsec:dynamic} (Lines $8-11$). After traversing all points of all frames, we compute the mean background velocity $\mathbf{v}_b$ and the mean velocity $\mathbf{v}_d^j$ for dynamic points on rigid object $j$ (Lines $17-18$). Then, background and dynamic points' neighborhoods are determined (Line $19$). Note that $d_0$ is a constant to reserve a minimum neighborhood range. We set $d_0 = 0.5$ m and $F=5$ in experiments. In the {\em for} loop, we superimpose the consecutive frames to the current frame (Line $7$). At last, we count the neighbors of each point in the current frame and mark ``spurious'' if the number of a point's neighbors is lower than the $5$\spth percentile (Lines $20-23$).




\begin{algorithm} 
\caption{Velocity-adaptive Spatial Stability Checking}
\label{alg:spatial_check}
\begin{algorithmic}[1]
  \STATE Input: Consecutive $F$ frames of point clouds, corresponding relative timestamp $\Delta t_i$ and rigid transformation $\mathbf{T}_i^0$ {\em w.r.t.} the current ($0$\spth) frame, $i = 1, 2, \cdots, F-1$
  \STATE Velocity set for dynamic points in rigid object $j$, $\mathcal{V}_d^j = \emptyset$, $j = 1, \cdots, G$; velocity set for background points $\mathcal{V}_b = \emptyset$
  \FOR {$i = 1 : F-1$} 
  	\STATE Compute the radar's velocity $\mathbf{v}_i^0$ by $\mathbf{T}_i^0$ and timestamps
  	\STATE $\mathcal{V}_b \leftarrow \mathbf{v}_i^0$
  	\FOR {each point $\mathbf{p}^i$ in $i$\spth frame}
  		\STATE Transform $\mathbf{p}^i$ to the current frame $\mathbf{p}_i^0$ by $\mathbf{T}_i^0$
  		\IF {$\mathbf{p}^i$ is a dynamic point in rigid object $j$}
  		  \STATE Compute this point's velocity $\mathbf{v}_p^j = \frac{\Delta\mathbf{d}}{\Delta t_i}$
  		  \STATE $\mathbf{v}_d^j = \mathbf{v}_i^0 - \mathbf{v}_p^j$
  		\ENDIF
  	 \ENDFOR
  	 \FOR {$j = 1 : G$}
  	 	\STATE $\mathcal{V}_d^j \leftarrow \mathbf{v}_d^j$
  	 \ENDFOR
  \ENDFOR
  \STATE Compute the time span of all frames $\Delta t = \sum_{i=1}^{F-1}\Delta t_i$
  \STATE Compute mean background velocity $\mathbf{v}_b = \text{mean}\{\mathcal{V}_b\}$ and mean velocity $\mathbf{v}_d^j = \text{mean}\{\mathcal{V}_d^j\}$ for all dynamic objects
  \STATE Background points' neighborhood range $d_b = \max\{d_0, \frac{1}{2}\|\mathbf{v}_b\|\Delta t\}$; dynamic points in object $j$ have neighborhood range $d_j = \max\{d_0, \frac{1}{2}\|\mathbf{v}_d^j\|\Delta t\}$
  \FOR {each point $p$ in $0$\spth frame} 
  	\STATE Count neighbors of point $p$ as $N_p$
  \ENDFOR
  \STATE Mark point $p$ in $0$\spth frame ``spurious'' if $N_p$ is lower than the $5$\spth percentile.
\end{algorithmic}
\end{algorithm}

\section{System Implementation and Evaluation}
\label{sec:evaluation}
\subsection{Implementation and Experimental Setup}
Currently, only the RPDNet dataset~\cite{rpdnet} provides raw radar RDMs with, however, LiDAR point clouds. No public dataset provides radar RDMs, monocular images, and IMU measurements. Therefore, we collect our dataset through a customized platform.

{\bf Hardware and dataset.} As shown in the left picture in Fig.~\ref{fig:pipeline}, the customized platform includes a 4D mmWave radar, a camera, and an IMU. The 4D radar is cascaded by four TI AWR2243, which contains $12$ transmitting antennas and $16$ receiving antennas. We use the Intel Realsense D435 to provide depth images and monocular images. We obtain the ground-truth 3D point cloud of visual features transformed from the depth images. We use the high-precision IMU integrated in CUAV v5+ flight controller to provide accelerations and angular velocities. All sensor data are timestamped and managed by the Intel NUC 11TNKi5 running Ubuntu 20.04 with robot operating system (ROS). At last, we transmit the dataset into a server for training. 

{\bf Dataset and training.} Our dataset includes $187200$ IMU sequences and $62400$ pairs of camera images and radar range-doppler matrices in indoor and outdoor scenes. We collect the outdoor data along a city road outside the campus, covering a duration over $2080$ seconds in a total distance of $5720$ meters. The indoor data are collected in our lab and offices. Across all our data, we use $2288$ meters long trajectories with $24960$ pairs of camera and radar data for training. 
During training, we take all sensor measurements. In the testing, we only input radar RDMs. 

{\bf Ground-truth point cloud.} We obtain the ground-truth point clouds from stereo images provided by Intel Realsense D435. Specifically, we set the same resolution for monocular images and stereo images. Then we first track scale-invariant feature transform features~\cite{lowe1999object} and calculate their 3D points from the corresponding pixels in the stereo depth maps 



{\bf Baselines.} We compare with Cheng~\et~\cite{cheng2022novel}, CA-CFAR, and OS-CFAR~\cite{richards2014fundamentals}. Cheng~\et~\cite{cheng2022novel} represents the SOTA 3D radar point cloud generation. CA-CFAR (Cell Averaging Constant False Alarm Rate) and OS-CFAR (Order Statistic CFAR) are two common techniques for target detection and estimation while maintaining a constant false alarm rate.


\subsection{Performance Evaluation}
{\em 1) Point cloud enhancement.} We would like to see if the enhanced point cloud is similar to the ground-truth point cloud. We use three metrics to evaluate the similarity: RPCDL~\cite{cheng2022novel}, Chamfer distance~\cite{chamfer}, and modified Hausdorff distance~\cite{hausdorff}. A higher RPCDL indicates that the radar point clouds are closer to the ground-truth point clouds. Chamfer and modified Hausdorff distances find the nearest neighbor for each point in one point cloud to another and takes the mean and median of all these distance respectively.

\begin{figure}[t!]
    \centering
    \shortstack{
            \includegraphics[width=0.23\textwidth]{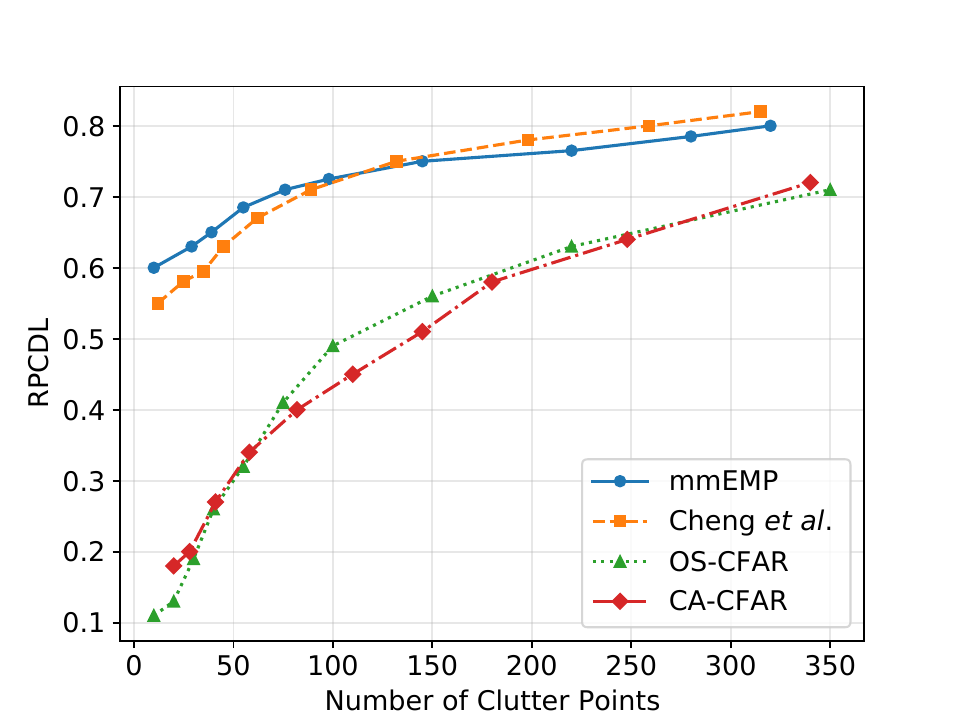}\\
            {\footnotesize (a) RPCDL}
    }
    \shortstack{
            \includegraphics[width=0.23\textwidth]{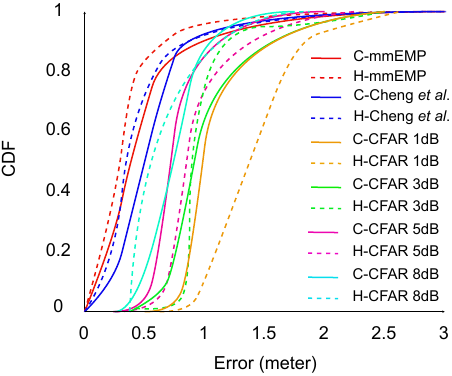}\\
            {\footnotesize (b) Chamfer and M. Hausd.}
    }
    \caption{Point cloud similarity.}
    \label{fig:rpcdl}
    \vspace{-4mm}
\end{figure}
Fig.~\ref{fig:rpcdl}(a) shows the relationship between the number of clutter points and the RPCDL. A radar clutter point is defined as the point who fails to find any stereo-generated point within a distance threshold $\delta$. In our experiments, $\delta = 1$ m. It can be seen that mmEMP performs similar to Cheng~\et~\cite{cheng2022novel}, which enhances point clouds with a costly LiDAR. When the number of clutter points is about $50$, mmEMP generates more than two times denser point clouds than the ones generated by OS-CFAR and CA-CFAR. Fig.~\ref{fig:rpcdl}(b) shows that mmEMP achieves the Chamfer and modified Hausdorff distances with the median errors of $0.38$ m and $0.30$ m, respectively, being close to Cheng~\et~\cite{cheng2022novel} with the median errors of $0.50$ m and $0.34$ m. This figure evaluates the CA-CFAR with threshold from $1$ dB to $8$ dB. The increase of the threshold decreases the point cloud density. Compare with CFAR methods, mmEMP is much more similar to the ground-truth point cloud. 



\begin{figure}[t!]
  \centering
  \includegraphics[width=3in]{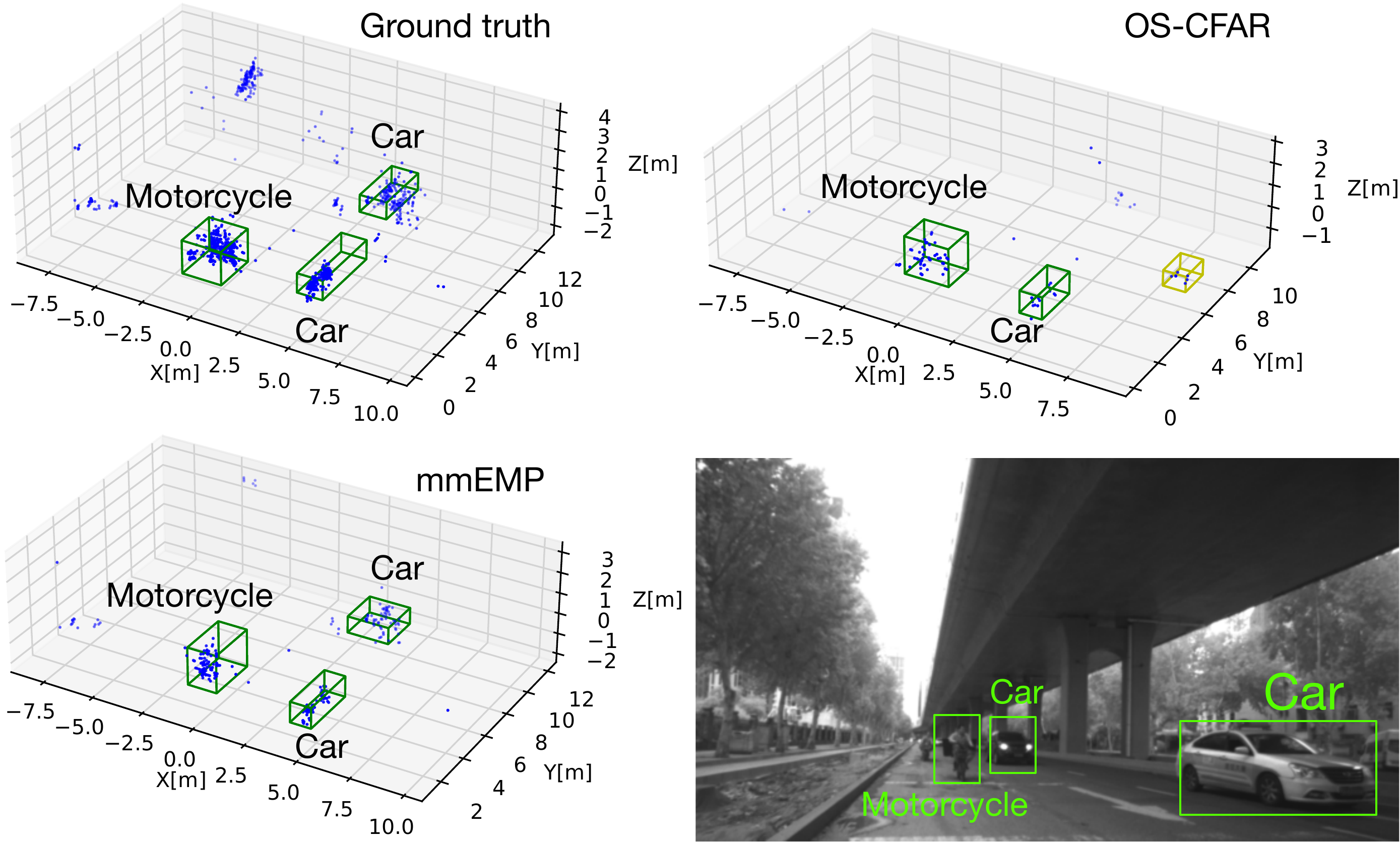}
  \caption{Object detection from ground-truth point clouds and radar point clouds generated by OS-CFAR and mmEMP.}
  \label{fig:object_detection}
\end{figure}

\begin{table}[t!]
	\centering
	\caption{Ablation Study for mmEMP Modules}
	\label{tab:ablation}
	\begin{tabular}{p{8mm} p{8mm} p{8mm} p{10mm} p{14mm}}
		\hline
		\textbf{DVIR$^1$} & \textbf{PR$^2$} & \textbf{RPCDL$\uparrow$} & Chamfer (m)$\downarrow$ & mod. Hausdorff (m)$\downarrow$\\
		\hline
		 &  & 59.07 & 0.342 & 0.387\\
		\checkmark & & 74.39 & 0.233 & 0.286 \\
		 & \checkmark & 68.68 & 0.258 & 0.312 \\
		\checkmark & \checkmark & {\bf 78.64} & {\bf 0.192} & {\bf 0.248} \\
		\hline
	\end{tabular}
	\begin{tabular}{l}
		$^1$DVIR denotes the dynamic VI 3D reconstruction module. \\
		$^2$PR denotes the point refinement module.
	\end{tabular}
\end{table}

{\em 2) Ablation study.} We evaluate the effectiveness of each module through ablation experiments, as shown in Table~\ref{tab:ablation}. Without any proposed module, we input the conventional VI 3D reconstruction~\cite{qin2018vins} into Cheng~\et~\cite{cheng2022novel}. The three metrics are the worst because dynamic visual features distort the 3D perception and spurious points bring fake objects. The results show that our dynamic 3D reconstruction leads to the biggest performance gain (refer to $2^{\text{nd}}$ row). $3^{\text{rd}}$ row shows the performance gain provided by the point cloud refinement. 

\begin{figure}[t]
	\centering
	\begin{minipage}[b]{0.19\textwidth}\centering
		\center
		\includegraphics[width=1\textwidth]{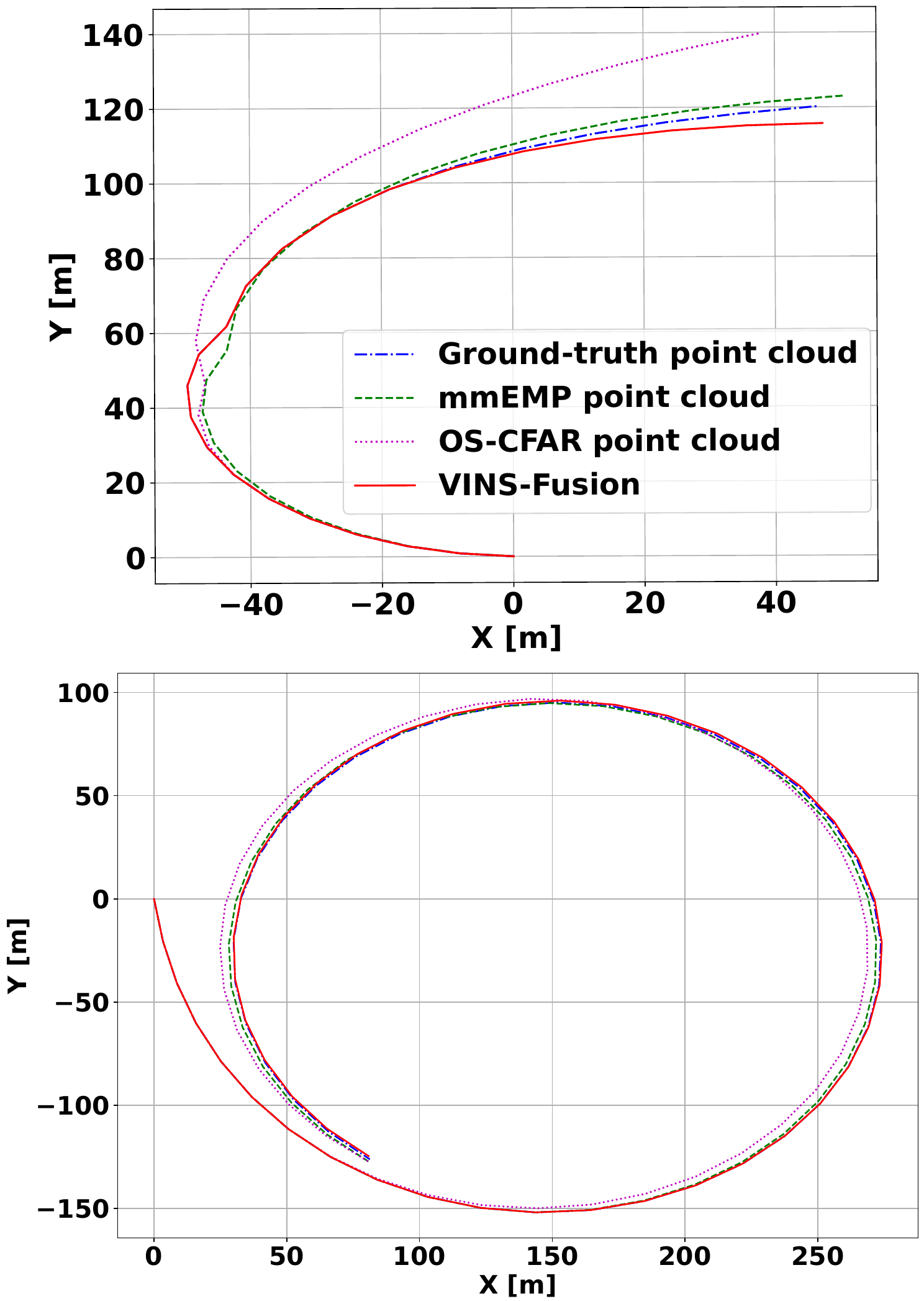}\vspace{-0.3cm}
		\caption{Localization using ground-truth and radar point clouds from OS-CFAR and mmEMP.} 
		\label{fig:localization}
	\end{minipage}
	\hspace{0.1cm}
	\begin{minipage}[b]{0.27\textwidth}\centering
		\center
		\includegraphics[width=1\textwidth]{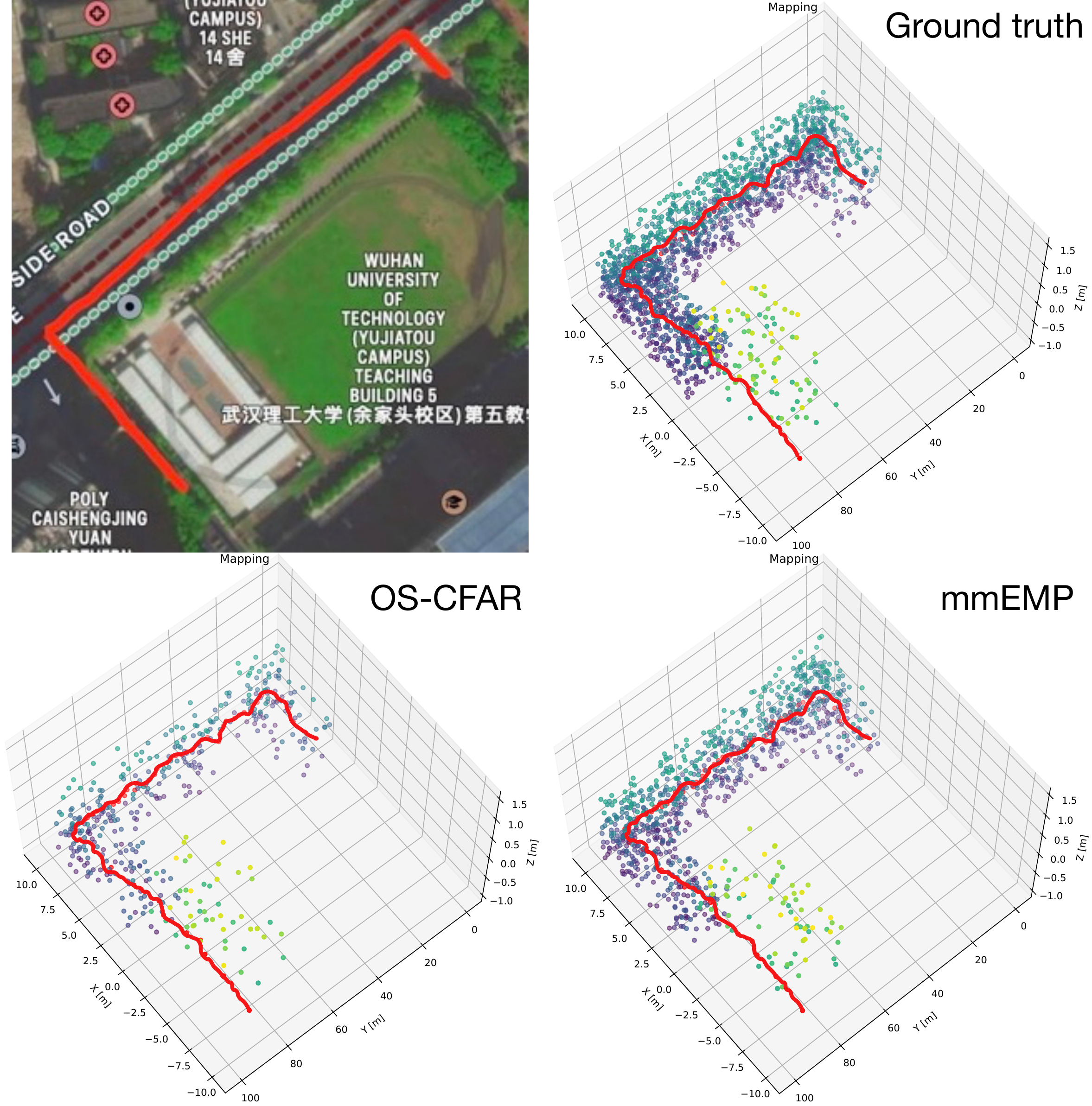}\vspace{-0.3cm}
		\caption{Mapping using ground-truth point clouds and radar point clouds from OS-CFAR and mmEMP.} 
		\label{fig:mapping}
	\end{minipage}
	\vspace{-0.4cm}
\end{figure}

{\em 3) Application on subtasks.} We demonstrate the effectiveness of mmEMP enhanced point clouds by object detection, localization, and mapping.
Fig.~\ref{fig:object_detection} shows that the spurious points in OS-CFAR lead to a false alarm detection, \ie, the yellow box. mmEMP generates a much denser point cloud so as to produce accurate object bounding boxes. 
Fig.~\ref{fig:localization} shows the localization qualitative results of two trajectories in the outdoor scene. The ground-truth odometry is obtained by VINS-Fusion~\cite{qin2019a} with GNSS readings to prevent the temporal drift of VINS-Mono~\cite{qin2018vins} without loop closure. The localization results with different point clouds are from the Kabsch algorithm in our neural network pipeline (refer to Fig.~\ref{fig:pipeline}). The mean localization errors {\em w.r.t.} the upper trajectory of using point clouds from the ground truth, mmEMP, and OS-CFAR are $1.38$ m, $2.19$ m, and $5.62$ m, respectively. For the bottom trajectory, the errors are $1.43$ m, $2.04$ m, and $5.66$ m, respectively. Fig.~\ref{fig:mapping} shows the mapping qualitative results. The mapping errors in Chamfer distance of using point clouds from mmEMP and OS-CFAR are $2.51$ m and $3.06$ m, respectively.

\section{Conclusion}
\label{sec:conclusion}
mmEMP creates dense point clouds from a single-chip mmWave radar to support autonomous driving through a supervised learning pipeline. We use the supervision from a camera-IMU sensor suite, which is low-cost and commonly available in commercial vehicles, enabling crowdsourcing training data. mmEMP overcomes the challenges arising from dynamic visual features and spurious radar points with a dynamic 3D reconstruction algorithm and a neural network design. We collect a large dataset of image-radar pairs with IMU sequences and conduct experiments in indoor and outdoor scenes. The results show that mmEMP generates denser point clouds without spurious points and thus contributes more to the perceptional tasks, including object detection, localization, and mapping. In the future, we will explore the potential of the enhanced radar data by fusing it with other sensors, \eg, event-based camera and WiFi beamforming reports, to build more robust robot perceptional systems at high speeds and in human-robot interactions.

\section*{Acknowledgment}
This work was supported in part by the National Natural Science Foundation of China (NSFC) under grant No. 52031009, in part by the Natural Science Foundation of Hubei Province, China, under grant No. 2021CFA001, in part by the NSFC under grant No. 62272098, No. 62372265, and No. 62302254, in part by the National Key Research Plan under grant No. 2021YFB2900100.

\bibliography{root.bib}
\bibliographystyle{IEEEtran}
\balance

\end{document}